\RequirePackage[2020-02-02]{latexrelease}
\documentclass{clv3}
\usepackage{comment}
\usepackage{hyperref}
\usepackage{xcolor}
\definecolor{darkblue}{rgb}{0, 0, 0.5}
\hypersetup{colorlinks=true,citecolor=darkblue, linkcolor=darkblue, urlcolor=darkblue}

\begin{document}
\issue{1}{1}{2022}

 \dochead{The Pitfalls of Defining  Hallucination}

\runningtitle{Defining Hallucination}

\runningauthor{Kees van Deemter}

\author{Kees van Deemter.}
\affil{Utrecht University}

\maketitle

\begin{abstract} 
Despite impressive  advances in  Natural Language Generation (NLG) and Large Language Models (LLMs),   researchers are still  unclear about important aspects of NLG evaluation. To substantiate this claim, I examine current classifications of    hallucination  and omission 
in Data-text NLG, and I propose a logic-based synthesis of these classfications.  I conclude by highlighting some remaining limitations of all current thinking about hallucination and by discussing implications for LLMs.
 \end{abstract}

\section{Introduction: evaluating the veracity of a text}

When computers produce text,   the quality of  the   texts  is of paramount concern. For this reason, a substantial body of work across Natural Language Generation (NLG) focusses on evaluation of generated text. 
Evaluation can offer insight in various aspects of the quality of a generated text; indirectly, by looking at a range of generated texts,  it can  tell us how well a given NLG technique works.

A key family of quality criteria, for most text {\em genres} at least, centers around what might be called the veracity  of the text. I will take the view that  assessing  the veracity of a text  means assessing whether the text ``speaks the truth, the whole truth, and nothing but the truth". Veracity is crucial: if a text is lacking in veracity, then  any other virtues  that it may possess -- such as fluency and clarity for example -- can only  amplify the risks that arise when the text is read    (cf., \citet{crothers2023machine}), because a well-written falsehood is more likely to be taken seriously than a badly written one.

To obtain a clear perspective, 
I will focus on traditional NLG tasks first, whose aim is to convert a structured input into a text. I will argue that, even with such a relatively simple task,  when we  assess the veracity of a text, we do not quite know what we are doing. 
I will  highlight some  flaws in current analyses of hallucination, and I will  offer a synthesis that does not suffer from these flaws. After that, I will argue that all current analyses still suffer from some important limitations, and I will  discuss implications for more complicated  tasks, as addressed by Large Language Models (LLMs).

\section{Existing analyses  of veracity}

When rule-based NLG systems are assessed, veracity is sometimes taken for granted; 
 with the ascent of neural architectures, researchers in various areas of NLG have emphasized  that this is often not justified \cite{vinyals2015neural,koehn2017six,rohrbach2018object,maynez2020faithfulness,duvsek2020evaluating,raunak2021curious}. Let's look briefly at three  attempts to analyse what the problem is and what forms it can take. Since these issues come  into focus most sharply in relation to Data-text NLG -- where the input to the generator is structured data, not text -- we focus  on that type of NLG; other types of NLG will be discussed in section 5.

One clear-headed, but coarse-grained, attempt came from  \citet{duvsek2020evaluating}, who  discussed Data-text NLG systems whose input was a set of atomic statements and whose output was a verbalization of each of these statements.\footnote{In other words,   the decision of ``What to say" (as opposed to ``How to say it"), also known as Content Determination, has already taken place  (see e.g. \citet{gatt2018survey}, section 3).} 
For some specified individual $x$,  the input could be $Type(x) = Restaurant \wedge Food(x)= Italian \wedge Price(x)=Low$, and the output ``$x$ is an affordable Italian restaurant". The authors highlighted  two possible problems, namely  hallucination  and  omission. 
For them, {\em hallucination} occurs when the output does not follow  from the input; {\em omission} happens when the input does not follow from the output. Thus, if the generator produces ``$x$ is an affordable {\em veggie} Italian restaurant", then the information that $Style(x)= vegetarian$ is hallucinated; if it produces ``$x$ is an Italian restaurant", then that involves an {\em omission}, because the information that $Price(x)= Low$ is omitted. Note that, by relying on the ``follows from" relation, Dusek \& Kasner's is, at heart, a {\em logical} analysis, where the input is  a formal meaning representation formula and the output is  natural language text. The authors phrase their idea in terms of Natural Language Inference (NLI),  assuming a version of NLI that ``crosses formats",  relating formulas and text (see section 4).   

A more finegrained analysis  of hallucination was offered by 
\citet{ji2023survey}, who distinguished between {\em intrinsic hallucination} and {\em extrinsic hallucination}: the former is ``output that contradicts the source”; the latter is ``output that can neither be supported nor contradicted by the source”. This analysis, whose core concepts (contradict, supported) appear likewise, though more implicitly than in the previous case,  to be rooted in logic, was applied to a number of  NLG areas,  including Data-text NLG; notably, omission of information was not discussed.

A third strand of research tries to classify all kinds of errors that are found in computer-generated texts   \cite{thomson2020gold,moramarco2022human,van2023barriers}. Since these errors include  hallucination, it will be instructive to  examine one such account, namely \citet{thomson2020gold}, who distinguished between outputs that contain an incorrect number, an incorrect named entity, an incorrect word (that is neither an incorrect number nor an incorrectly named entity), ``non-checkable" information, 
context errors, and other  kinds of errors.

 To see how these three analyses compare, let's look at some outputs that may be generated on he basis of a given input. 
 Suppose the input is, once again, $Type(x) = Restaurant \wedge Food(x)= Italian \wedge Price(x)=Low$. 
The outputs are as follows:
\begin{itemize}
\itemsep0em 
    \item[Ex1.] ``$x$ is an Italian restaurant."\\
    {\bf D\&K}: Omission. {\bf Ji }: n.a. {\bf T \& R}: n.a.
     \item[Ex2.] ``$x$ is  an affordable veggie Italian restaurant".\\
       {\bf D\& K}: Hallucination. {\bf Ji }: Extrinsic hall.  {\bf T\&R}: word error.
    \item[Ex3.] ``$x$ is  a veggie restaurant".\footnote{Even with this simple input, it is not always obvious  how  the proposed definitions should be applied. For example, Thomson and Reiter's scheme might  alternatively be used to categorize Ex2 as non-checkable, and Ex4 as involving an incorrectly named entity (namely Norway). }  \\  
    {\bf D\&K}: Hallucination and Omission. {\bf Ji } : Extrinsic hall.  {\bf T\&R}: word error    
     \item[Ex4.] ``$x$ is  an affordable Norwegian restaurant".\\
    {\bf D\&K}: Hallucination and Omission. {\bf Ji }: Intrinsic hallucination. {\bf T\&R}: n.a.    
\end{itemize}

 Similar patterns emerge from different kinds of input. Suppose, for example, a numerical input for weather reporting specifies that the temperature is above $22^0$   Celsius. Now consider the  output ``The temperature is above 21 degrees". With this type of input, the definitions are not entirely clear cut (e.g., because it can be difficult to isolate the precise part of the output that is at fault), but I take it that both {\bf D\&K} and {\bf Ji} would analyse this output  exactly like  (Ex1) above;   ``The temperature is above 23" would be analysed  like  (Ex2); ``The temperature is below 25" would be analysed as Ex3, and ``The temperature is below 22" as Ex4. I take it that {\bf T\&R} would analyse each of these  outputs as an Incorrect Number error.

Clearly, these analyses differ  substantially from eachother, echoing the complaint in \citet{huidrom2022survey} that  classifications of  NLG errors tend to disagree with each other. Moreover, each analysis conflates two or more types of misinformation, each of which would pose different kinds of risks if they occurred in real life. Dusek \& Krasner's analysis, for example, does not distinguish between  (Ex3) and (Ex4), because all it can say about both situations  is that omission   and  hallucination occur (because the input does not imply the output and the output does not imply the input). Ji et al.  conflate (Ex2) and (Ex3), because both  ``can neither be supported nor contradicted by the source".  

If we wish  to use  concepts  such as hallucination  as a tool for evaluating the veracity of NLG models (using either human annotations or computational metrics or both),   and as a starting point for mitigating different types of generation errors, then  we need to go back to the drawing board.

\section{A synthesis of existing analyses }

Luckily,
a synthesis  of the analyses by Dusek and Kasner and  Ji et al.  is possible. To obtain a   systematic perspective that is applicable to domains of all kinds, let us step back and  ask what Logical Consequence (i.e., ``follows from", ``$\models$") relations can exist between input and output, assuming a classical logic. Assume, for now, that  neither the input nor the output to the generator is internally inconsistent.  It can be true or false that $input \models output$; likewise, it can be true or false that $output \models input$; furthermore, if $input \not\models output$ and $output \not\models input$,  it matters whether $input \models \neg output$ (as in Ji et al.'s intrinsic hallucination, where output contradicts input), splitting error type 3 in two:\footnote{In  (1), $input \models \neg output$ would imply that the  $input$ is inconsistent (because also $input \models  output$). In (2), $output \models \neg input$ would imply that the $output$ is inconsistent (because also $output \models  input$).}
\begin{itemize}
\itemsep0em 
    \item[0.] $input \models output$ and $output \models input$. 
 (Input and output are well matched.)
    \item[1.] $input \models output$ but $output \not\models input$. (Output is too weak.) 
     \item[2.] $input \not\models output$ but $output \models input$. (Output is too strong.)  \item[3.] $input \not\models output$ and $output \not\models input$. (Neither follows from the other.)\\
      \begin{itemize}
    \itemsep0em  
\item[3a.]  (3) and $input \not\models \neg output$. (Input and output are logically independent of each other). E.g., 
``$x$ is  a veggie restaurant."
 \item[3b.]  (3) and $input \models \neg output$. (Input and output contradict each other.)\\
E.g., ``$x$ is  an affordable Norwegian restaurant."    \end{itemize} 
   \end{itemize}
If one also wishes to takes into account that outputs can be self-contradictory (as documented by \citet{moramarco2022human}, for example) or tautologous, the logical analysis can be pushed further by splitting  category 1  and  2 as well:
\footnote{In classical logic: a tautology follows from everything (1b); everything follows from a contradiction (2b).} 
\begin{itemize}
\itemsep0em 
    \item[1.] $input \models output$ but $output \not\models input$. (Output is too weak.)
    \begin{itemize}
    \itemsep0em  
\item[1a.] (1) and $ \not\models output$. (Output is not tautologous; this is the normal case). 
E.g., ``$x$ is an Italian restaurant."
 \item[1b.]  (1) and $\models output$. (Output is  tautologous). E.g., ``$x$ is  Italian or not."
      \end{itemize}
     \item[2.] $input \not\models output$ but $output \models input$. (Output is too strong.)\\
     \begin{itemize}
    \itemsep0em  
\item[2a.] (2) and $ \not\models \neg output$. (Output is not  contradictory; this is the normal case). 
E.g., ``$x$ is  an affordable veggie Italian restaurant." 
 \item[2b.] (2) and  $\models \neg output$. (Output is  contradictory).\\ E.g., ``$x$ is a veggie steak restaurant."
      \end{itemize}
\end{itemize}

  It is easy to see how this new analysis applies to  other types of information, for example when the input is numerical. For instance, if  the input says the   temperature is between $20^0$ and $30^0$ Celsius, then an output that says ``The temperature is between  $25^0$ and $35^0$" would be logically independent of the input and hence inhabit category 3a. 
  
 Note that, for better or worse, the {\em truth} of the output, in the real world,  has not been a consideration: our core question has been whether the output of the generator matches the input. 

\section{Limitations of these analyses} 

This synthesis can underpin a computational metric that records how often each hallucination type occurs in a given (generated) text, provided the ``follows from"  relation can be computationally modeled using Natural Language Inference (NLI) (as proposed by \citet{duvsek2020evaluating}). %
 This presupposes that typos and incorrect   names should somehow be finessed (e.g., \citet{faille2021entity}),  
 which  may or may not be seen as affecting the veracity of a text.
  More problematically, it
 assumes that NLI is able to deal with ambiguous and vague text.  
 For instance, if the input specifies that the temperature is above $22^0$ Celsius, does it follow  that ``It's a warm day"?
Such issues  may well  be beyond the present state of the art  of NLI   \cite{liu2023we}.

Our analysis has some intrinsic limitations. It's best seen as the topic-independent  part of a full analysis, which disregards everything that's specific to a specific topic or application. When applied to an NLG system whose texts offer treatment advice to doctors, for example, then further distinctions will need to be made to determine whether an error is medically significant or not (cf., \citet{moramarco2022human}).

{\bf Pragmatic reasoning should always be applied.} 
When a multi-sentence text is generated,
each sentence should be given an interpretation appropriate for its context,\footnote{This echoes the idea of Context Errors,  a separate category in \citet{thomson2020gold} .} 
for instance with anaphoric pronouns resolved. Likewise, any pragmatic implicatures (in the style of \citet{grice1975logic}) of the sentence should be taken into account.
Similar remarks apply  to presupposition, irony, and metaphor,  which go beyond what is stated literally in a text, yet all of which can cause  hallucination and omission. 
For example, when the weather report for a sunny day speaks metaphorically of ``wall-to-wall sunshine", then a narrowly semantic analysis might mis-classify this  as a (3a-type) hallucination, but an analysis that understands metaphor should consider it to be truthful (i.e., well matched). In other words, measures should be taken to ensure that the ``follows from" relation is pragmatically, as well as semantically, aware. 

{\bf Even a domain-independent analysis could be driven further. } The classical ``follows from" relation cannot tell us everything one might want  to know about the relation between input and output. For a start, it  does not tell us anything about the {\em amount} of information that is added to or omitted from the input, only whether there exists such information. 
More subtly, suppose our input is, once again, $Type(x) = Restaurant \wedge Food(x)= Italian \wedge Price(x)=Low$. Then the output ``$x$ is a Norwegian  restaurant" 
falls into the same category as ``$x$ is an affordable Norwegian restaurant" 
even though, unlike the latter, this output {\em omits} some information (``affordable") from the output. 
And since our analysis is based on classical logic, it is unable to tell us what a text is {\em about}.  Assuming the same input as before, an output that says ``The cat is on the mat" would fall  into the same category as ``$x$ is a veggie restaurant", 
because both outputs are logically independent of the input; the fact that the sentence about the cat is also topically unrelated to the input is something that our analysis is unable to pick up. --- I take it that these limitations are acceptable once we are aware of their existence.
  
To assess whether an NLI-based implementation of our analysis matches the opinions of human judges, we are currently conducting a pilot study in which domain experts are asked to apply our analysis  to  commercial advertisements.   A challenge facing us, in this domain, is  situations in which the NLG output contains  information that is not present in the input,  
but where  the added information can nonetheless be inferred  with high probability. The question is, are these hallucinations or not? 
In other cases, the texts venture gratuitous information like ``This place has a family-friendly atmosphere". Although such information cannot be inferred from the input, we are encouraging annotators   to turn a blind eye 
because the falsity of the added  information would be difficult to  prove in a court of law (e.g., if a customer decided to complain about the atmosphere at this place). Note that this approach goes beyond  the proposal in section 3, because it asks not whether the output follows from the input, 
but whether the output is (or is likely to be) true in the real world. It is time that we examine this distinction more closely.   
 
\section{Veracity in Large Language Models and in other tasks than  Data-text NLG} 

 With the current popularity of LLM-driven Generative AI, 
 the question is being raised, not only by researchers but by society at large, what is the veracity of the texts that are generated by LLMs. These issues matter greatly because LLMs are starting to be employed in real-world contexts, by people who are no experts in NLP, and who may not always be aware how veracity may be compromised. Identifying the ways in which LLMs can go wrong is a necessary first step towards identifying real-world risks and technical mitigation strategies.
 
 Accordingly, a number of research groups have  started to investigate how veracity problems in LLMs can be classified and mitigated   \cite{zhang2023siren,huang2023survey}, discussing a wide range of interesting problems and potential solutions. veracity.
The question comes up whether, on top of these important endeavours, it might be possible to arrive at an analysis along the lines of section 3, 
which might reduce the risk that error categories might be overlooked or unclearly defined.

So, what happens when one tries to define the different ways in which  LLMs can be lacking in veracity?  LLMs have been employed to generate texts for a large variety of purposes. 
These differences are so substantial that veracity cannot mean  the same thing in all these cases.  In fact, for purposes of error classification, it does not matter what architecture (e.g., using LLMs or some other technique) is employed to perform a certain NLP task. But, contrary to what some writing in this area might lead one to expect,  it matters hugely {\em what that task is}; let's see why.

The analyses of veracity discussed in section 3 centered around the  relationship between the  input of the NLG system and its textual output, asking whether the output is too weak, or too strong, for its input, for instance. 
This perspective carries over almost {\em verbatim}  when the purpose of an LLM is to express all the information in the input truthfully, as in  \citet{lorandi2023data,yuan2023evaluating} for example. (Since LLMs operate on text, their generation  process starts by converting structured input into a linearized format, using mark-up  to convey logical structure (cf., \citet{harkous}), but I take this conversion step to be almost trivial.)  Consequently, all the distinctions that were made in section 3 apply to these LLM uses as well, including distinctions that are seldom made in the literature on LLMs, such as the distinction  between 1a and 1b, and the one between 3a and 3b. The perspective of section 3 may also carry over to some types of  Table-to-Text 
generation and Question Answering; it might also apply to  Machine Translation (e.g., \citet{dale2022detecting}), which  likewise hinges on an equivalence between input and output; the main difference with the type of classical NLG discussed in earlier sections is that where the latter generate text from structured input, the former generate text from text.

Some other uses of LLMs take up a  middle ground, where the perspective of  section 3 is  partially, but not wholly, applicable. When LLMs (or other NLP architectures, for that matter) are used for text summarization (e.g., \citet{pu2023summarization}) or  caption generation (e.g., \citet{rotstein2024fusecap}),  the requirement that $input \models output$   still makes sense (because a summary is meant to be faithful to the thing  that is summarized), but the requirement that $output \models input$  requires nontrivial modification  (because only the most important aspects on the input need to end up in the output summary). Our analysis is   applicable in full if and only if the task of the NLG system is to express all information in its input. 

Defining hallucination is even more problematic in open-ended LLM applications such as unrestricted Question Answering (see \citet{bubeck2023sparks,zhang2023siren,huang2023survey}
for plenty of examples) or essay writing (e.g., \citet{fitria2023artificial}) because, in such applications, it would be difficult to say {\em what input} (in our sense of that term) such an LLM is expressing.\footnote{If one wishes to view all the data that determine the LLM's response as input to the generator, this  implies that such an input  tends to be internally  inconsistent, necessitating the use of a para-consistent,  non-classical  logic (see e.g. \citet{paraconsistent}). Of course the use of a paraconsistent logic would  complicate the idea of seeing the output of a generator as ``following from" its input considerably.} As was pointed out by  \citet{zhang2023siren}, such systems suffer not only from  what these authors call  ``input-conflicting” information (where the LLM’s output deviates from the prompt provided by users), and from ``context-conflicting” information (where the output conflicts with information previously generated by the system), but also,  most important of all,  from ``fact-conflicting” information (where the output conflicts with world knowledge). Note that the output of such LLMs can still be assessed in terms of whether it is  tautologous (our type 1b) and whether it is contradictory (type 2b); testing for other kinds of fact-conflicting information, however, would require assessing whether the output is objectively true (replacing our earlier question of whether $input \models output$), and  possible also whether the output is informative enough for the application at hand (replacing the question of whether $output \models input$); in practice, of course, such assessments are sometimes difficult to make with certainty. In a political essay, for example, who is to say what is objectively true, 
or what it means for an essay to be informative enough?
 The Leibnizian idea that one can always answer such questions objectively is optimistic to say the least.

 \ \\
 But although this is treacherous territory, formal logic might once again help us on our way,  {\em via} a  logic  of  beliefs, desires and intentions (BDI)  \cite{bratman1987intention}. (See also \citet{van2020editors} for a broader inter-disciplinary perspective). BDI logics  
 add to our analytical arsenal because they allow us to reason about the inferences that a hearer will draw from an utterance, and to formally elucidate  notions such as lying and misleading \cite{sakama2010logical}. 
 Suppose, for example,  today's weather forecast does not mention a hurricane,  then listeners may infer that no hurricane will take place; if a hurricane hits them  nonetheless, they were misinformed by the forecast.\footnote{Using $q$ to abbreviate ``There will be a hurricane", and  $ C_{SH} p$ to say that the speaker (S)  communicates to the hearer (H) that $p$ , this can be expressed (slightly modifying  the formalism of \citet{sakama2010logical}   for current purposes) as the conjunction $\neg C_{SH} q$ {\em and} $B_H(q \rightarrow  C_{SH} q$)  {\em and} $\models q$.}  Omissions of this kind are, in the terminology of Sakama and colleagues,  ``withholding information"; they are important because   misinformation often hinges on  strategic omissions.  Another type of misleading highlighted by Sakama and colleagues is ``half truth":  an output is a half truth  if a  false proposition $r$ is not communicated directly, yet a truthful output is generated of which the hearer believes that $r$ follows from it.\footnote{In modified BDI notation,  with $p$ as the output,
 $ C_{SH} p$  {\em and}
  $\neg C_{SH} r$ {\em and} 
 $ B_H(p \rightarrow r)$ {\em and} $ \models p$ and $\not\models r$.} 
  Sakama et al's example is of a speaker bragging that he holds a permanent position at a company, without saying that the company is  almost bankrupt. 
  
  A BDI analysis does not by itself  tell us whether a given LLM-generated text withholds information, or tells a half truth; no existing NLI system would be able to tell us whether this is the case. On the other hand, BDI logic may give  us  a starting point for understanding  these matters. It would be interesting, for example, to try out whether annotators might be able to apply a version of  \citet{sakama2010logical}'s categories to essay texts; in this case, substantial levels of disagreement  between annotators are to be expected.

\section{Conclusion} 

 Natural Language Processing has made great strides, but our lack of understanding of 
 some of the most important issues in the evaluation of generated text should be cause for  humility, and even worry. To start addressing this problem, the NLP community should be prepared to do two unfashionable things: first, it should liaise with logicians;  after all, the latter have always focussed on concepts like truth and evidence. The synthesis offered in section 3, and the more speculative ideas in section 5, indicate the benefits that can be gained in this way. Secondly, work on NLG evaluation should pay more attention to the most "difficult" aspects of communication, including phenomena such as ambiguity and vagueness, which undermine our current understanding of logical inference (see e.g., \citet{KvD}), and including
 the crucial distinction between what a sentence   asserts  and what an utterance of the sentence 
 communicates {\em via} such mechanisms as implicature, presupposition, irony, and metaphor (see e.g., \citet{levinson1983pragmatics}). 
 
 Now that all of us are becoming  consumers of computer-generated text, the NLP community  disregards  such matters at everyone's peril.\\[2ex]

   \ \\[1ex] {\small {\bf Acknowledgment.} I am grateful for comments from reviewers; from people in Utrecht's NLP group;  and from Lasha Abzianidze, Guanyi Chen, Hans van Ditmarsch, and Ehud Reiter.}

\pagebreak

\starttwocolumn

 \bibliography{lastwords.bib}

\end{document}